\documentclass[11pt,a4paper]{article}

\usepackage{amsmath,amssymb,amsthm}
\usepackage{polyu-vclab}

\usepackage{lipsum}                
\usepackage{hyperref}       
\usepackage{url}            
\usepackage{booktabs}       

\usepackage{nicefrac}       
\usepackage{microtype}      
\usepackage{xcolor}         
\usepackage{colortbl}

\usepackage{threeparttable}
\usepackage{caption}
\usepackage{graphicx}
\usepackage[table]{xcolor}
\usepackage{multirow}
\usepackage{array}
\usepackage{wrapfig}
\usepackage{overpic}
\usepackage{subfig}
\usepackage{tcolorbox}
\usepackage{enumitem}
\usepackage{algorithm}
\usepackage{algpseudocode}
\usepackage{setspace}
\usepackage{makecell}
\newcommand{\figref}[1]{Fig.~\ref{#1}}
\newcommand{\tabref}[1]{Tab.~\ref{#1}}

\renewcommand{\algref}[1]{Alg.~\ref{#1}}
\renewcommand{\eqref}[1]{Eqn.~(\ref{#1})}
\newcommand{\appnref}[1]{\textbf{Appendix~\ref{#1}}}
\newcommand{\myPara}[1]{\vspace{.05in}\noindent\textbf{#1}}
\usepackage[numbers,sort&compress]{natbib}
\setEyebrow{PolyU VCLab\,\textbullet\,Preprint 2026}

\setReportTag{Visual Computing Lab\,\textperiodcentered\,The Hong Kong Polytechnic University}

\papertitle{HaloBlock: Faster and Better Decoding for Diffusion Large Language Models}

\paperauthors{%
  Zhaohui Zheng\affilmark{1}\quad
  Chenhang He\affilmark{1}\quad
  Jinrui Zhang\affilmark{1}\quad
  Ming-Ming Cheng\affilmark{2}\quad
  Lei Zhang\correspondmark\affilmark{1}%
}

\paperaffil{%
  \affilmark{1}\,The Hong Kong Polytechnic University\quad
  \affilmark{2}\, College of Computer Science, Nankai University
}

\papernotes{%
  \correspondmark\,Corresponding author (\href{mailto:cslzhang@comp.polyu.edu.hk}{cslzhang@comp.polyu.edu.hk}).%
}

\paperbadges{%
  \vclabbadgesolid{Code}{https://github.com/PolyU-VCLab/HaloBlock}\;%
}

\begin{document}

\maketitleVCLab

\begin{vclabAbstract}
\noindent\textbf{Abstract.}\;
Diffusion large language models (dLLMs) have attracted increasing research attention due to their bidirectional attention and arbitrary order decoding.
However, there are two issues in dLLMs inference: the decoding difficulty is dominated by the tokens on the right side of the block, and the block-wise decoding inherently suffers from boundary contextual discontinuity.
To tackle these limitations, we propose HaloBlock, a difficulty‑aware adaptive halo mechanism for dLLM inference.
We first leverage bidirectional cumulative entropy to characterize block decoding difficulty.
Then, based on the bidirectional decoding difficulty, we adaptively allocate bidirectional halo regions on both sides of each block to mitigate boundary contextual discontinuity.
The difficulty‑aware design of HaloBlock smooths cross‑block context propagation and lowers decoding difficulty for hard tokens, yielding better generation quality and faster convergence.
Evaluated on LLaDA‑8B‑Instruct, Dream‑7B‑Instruct and LLaDA‑1.5 across mathematical reasoning and code generation benchmarks, HaloBlock demonstrates consistent accuracy improvements and higher inference efficiency.  Source codes are publicly available at \textcolor{blue}{https://github.com/PolyU-VCLab/HaloBlock}.

\end{vclabAbstract}


\section{Introduction}

Diffusion large language models (dLLMs) \citep{llada,dream7b,mmada} have recently emerged as a promising alternative to the classic autoregressive (AR) LLMs, attracting rapidly growing research attention due to their parallelism.
Equipped with inherent bidirectional attention and arbitrary-order decoding, dLLMs exhibit remarkable performance on long-range understanding and controllable editing tasks \citep{lee2025editext,li2026lavida}.
Despite their architectural advantages, dLLMs suffer from substantially high inference overhead and the convergence difficulty in decoding text.
Pioneering works \citep{block-diffusion,fast-dllm} have established block diffusion for semi-autoregressive decoding and key-value (KV) cache for inference acceleration, both of which have become the standard components in modern dLLM inference.

\begin{figure*}[!t]
	\centering
	\setlength{\tabcolsep}{1pt}
	\setlength{\abovecaptionskip}{3pt}
	\includegraphics[width=1\textwidth]{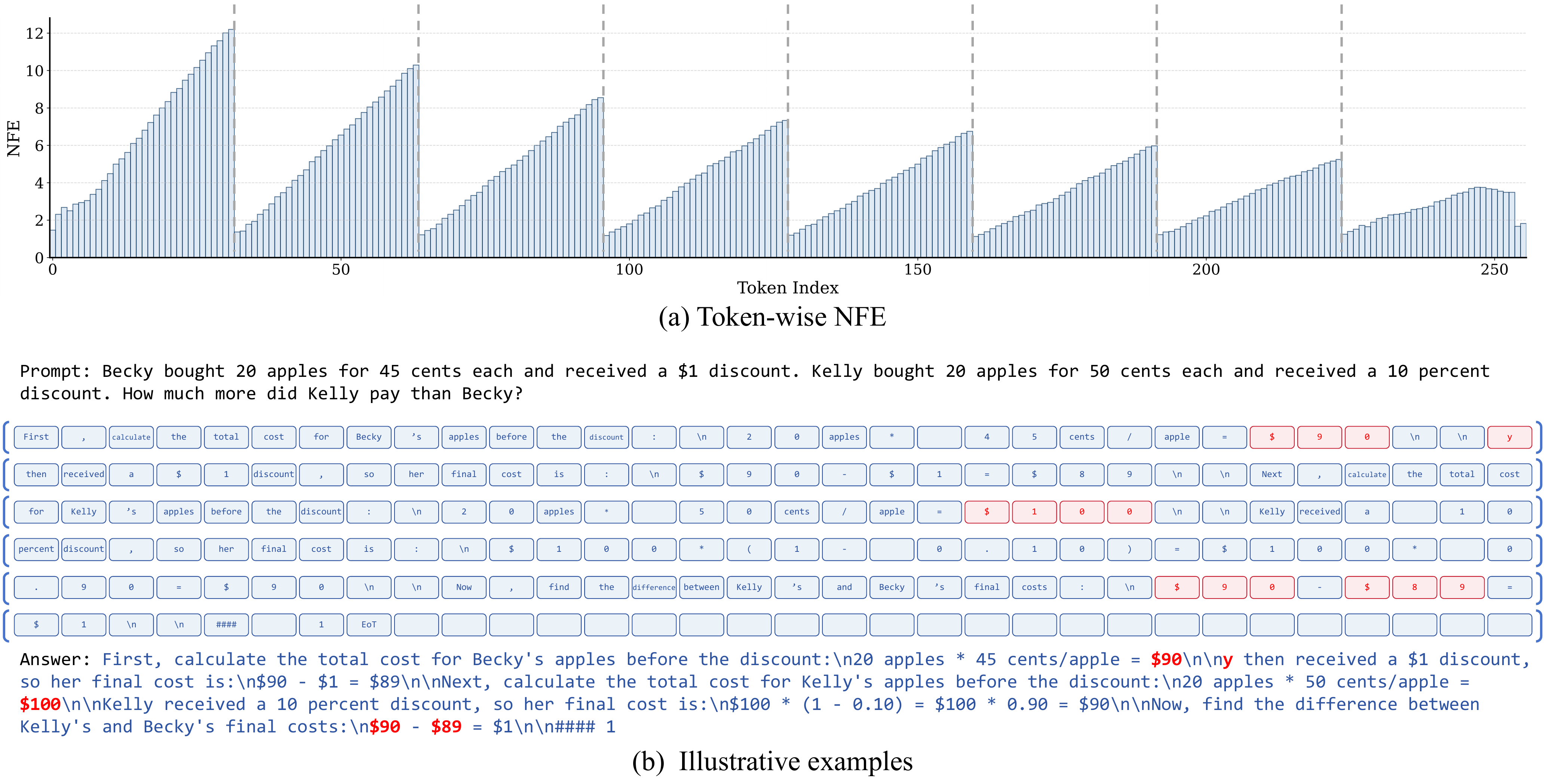}
    \captionsetup{font={small}}
	\caption{Illustration of the inference for dLLMs using LLaDA-8B-Instruct. (a) Token-wise Number of Function Evaluations (NFE). (b) An example of mathematical reasoning. The reasoning errors are highlighted in \textbf{\textcolor{red}{red bold}}. We see the left-dominance effect in block-wise decoding. Zoom in for a better view.}
	\label{fig:example}
 \vspace{0pt}
\end{figure*}

While block-wise decoding successfully alleviates the difficulty of global decoding, it introduces new challenges.
The first is the inherent imbalanced difficulty in intra-block decoding.
\figref{fig:example}(a) shows the sawtooth‑shaped variation in per-token NFE across token index.
The left-side tokens receive stronger support from preceding context, which significantly reduces their decoding difficulty, whereas tokens on the right benefit far less from contextual cues and thus face substantially higher decoding difficulty (a failure example is shown in \figref{fig:example}(b)).
This position-dependent convergence heterogeneity is broadly consistent with the observations in adaptive parallel decoding literature \citep{APD,adablock}, but its intra-block form in block-wise dLLM inference has not been identified or formalized.
Furthermore, as the total forward passes per block are dictated by the hardest tokens, this uneven difficulty distribution becomes a non-trivial issue on inference efficiency.
We refer to this imbalanced difficulty in intra-block decoding as left-dominance effect.
Second, a less discussed issue is the contextual discontinuity at block boundaries.
This discontinuity arises primarily from the asynchronized KV cache update across blocks, a fundamental limitation of block-wise decoding designs \citep{fast-dllm}.
During intra-block decoding iterations, the KV cache outside the current block is held constant.
This forces intra-block to interact with stale context and thus undermines cross-block semantic coherence.

To address the above challenges, we propose HaloBlock, a token-difficulty-aware decoding scheme for dLLMs.
Specifically, we first employ bidirectional cumulative entropy as a token-level metric to represent the token decoding difficulty.
Based on the estimated difficulty, we adaptively allocate halo regions to both ends of each decoding block.
The left halo consists of already decoded tokens from preceding blocks, while the right halo is composed of masked placeholder tokens.
With bidirectional attention, these halo regions effectively expand the receptive field of the current block and refresh the contextual information at boundaries, alleviating the boundary contextual discontinuity caused by asynchronous KV cache updates.
As a result, HaloBlock retains sufficient contextual coverage for high-uncertainty regions, which eases the decoding difficulty and achieves faster decoding convergence.
Our contributions are as follows:

\begin{itemize}
    \item We reveal the key structural limitations, i.e., left-dominance effect and boundary contextual discontinuity, in block-wise decoding of dLLMs. 
    \item We present HaloBlock, a training-free difficulty-aware decoding framework that adaptively allocates halos to each block, which effectively reduces the decoding difficulty and smooths the context generation at block boundaries.
    \item Experiments on representative dLLM, including LLaDA-8B-Instruct, Dream-7B-Instruct, and LLaDA-1.5, verify the effectiveness of HaloBlock, which yields clear improvements in generation accuracy and even achieves a modest speedup in inference.
\end{itemize}

\section{Related Work}

The predominant auto-regressive (AR) language models \citep{GPT-3,GPT-4,Llama1,Llama2,qwen} have long been used in Natural Language Processing.
Recently, diffusion large language models (dLLMs) \citep{sahoo2024simple,llada,dream7b} emerge as a promising alternatives to AR models.
Unlike AR models that are constrained to strict left-to-right generation, dLLMs are equipped with bidirectional attention and support arbitrary-order decoding, granting them unique strengths in long-range context understanding and controllable text editing \citep{lee2025editext,li2026lavida}.
There are mainly two directions to further improve dLLMs: to accelerate the inference speed and to improve the generation quality.
Block-diffusion \citep{block-diffusion} established a semi-AR model by introducing block discrete denoising diffusion.
Fast dLLM \citep{fast-dllm} introduced prefix cache and dual cache to enable KV cache reusing for the full bidirectional attention dLLMs.

Beyond the KV cache improvements \citep{dkv-cache,dllm-cache}, LLaDA \citep{llada} firstly demonstrated that a strong reasoning performance can be achieved by training a large-scale dLLM from scratch.
For generation quality, it has been found in AdaBlock \citep{adablock} that a reasonable block size is related to the semantic segment of the sequence: better alignment, better reasoning accuracy.
Unmasking or re-masking usually requires multiple hyper-parameters and meticulous selection pipeline to converge \citep{WINO,Hierarchy}.
In addition, researchers have attempted to add some assistant blocks, tokens, or masks to calibrate, supplement, and expand the decoding range \citep{li2026beyond,dpad}.

Apart from the above training-free methods, there are some training approaches to enhance the parallel decoding for dLLMs as well.
DParallel \citep{dParallel} performs self-distillation on pre-trained diffusion LLMs to encourage the models to decode more tokens at a time.
DMPO \citep{DMPO} studies a distribution matching policy optimization framework tailored for reinforcement learning fine-tuning of dLLMs.
D2F \citep{D2F} introduces discrete diffusion forcing to simultaneously enable accurate block-level KV cache and parallel decoding across multiple blocks through a causally-restricted attention architecture.
D3LLM \citep{d3llm} employs pseudo-trajectory distillation to guide dLLM inference.
Among all the aforementioned methods, a prevailing belief is that block-wise decoding and KV cache have become the standard components of dLLMs to improve decoding quality and speed, respectively.

\section{Methodology}

We start with a standard conditional text generation problem.
A sequence $X$ to be generated of length $T_x$ is initialized by a series of $[\mathtt{MASK}]$ tokens.
Together with the prompt $P$ of length $T_p$, we form the input sequence $S=[P,X]$. $X$ is decoded iteratively from the $[\mathtt{MASK}]$ tokens to the sampled tokens in the vocabulary space $\mathcal{V}$.
Block-wise decoding, which was first introduced by \citep{block-diffusion}, partitions the sequence $X$ into a series of sub-sequences $X_i$, each of which consists of a mount of tokens with length $t$ as the block size.
The decoding algorithm is performed on the blocks one by one from left to right.
A static KV cache $Y$ is used to interact with $X$ for inference acceleration.

\begin{figure*}[!t]
	\centering
	\setlength{\tabcolsep}{1pt}
	\setlength{\abovecaptionskip}{3pt}
	\includegraphics[width=1\textwidth]{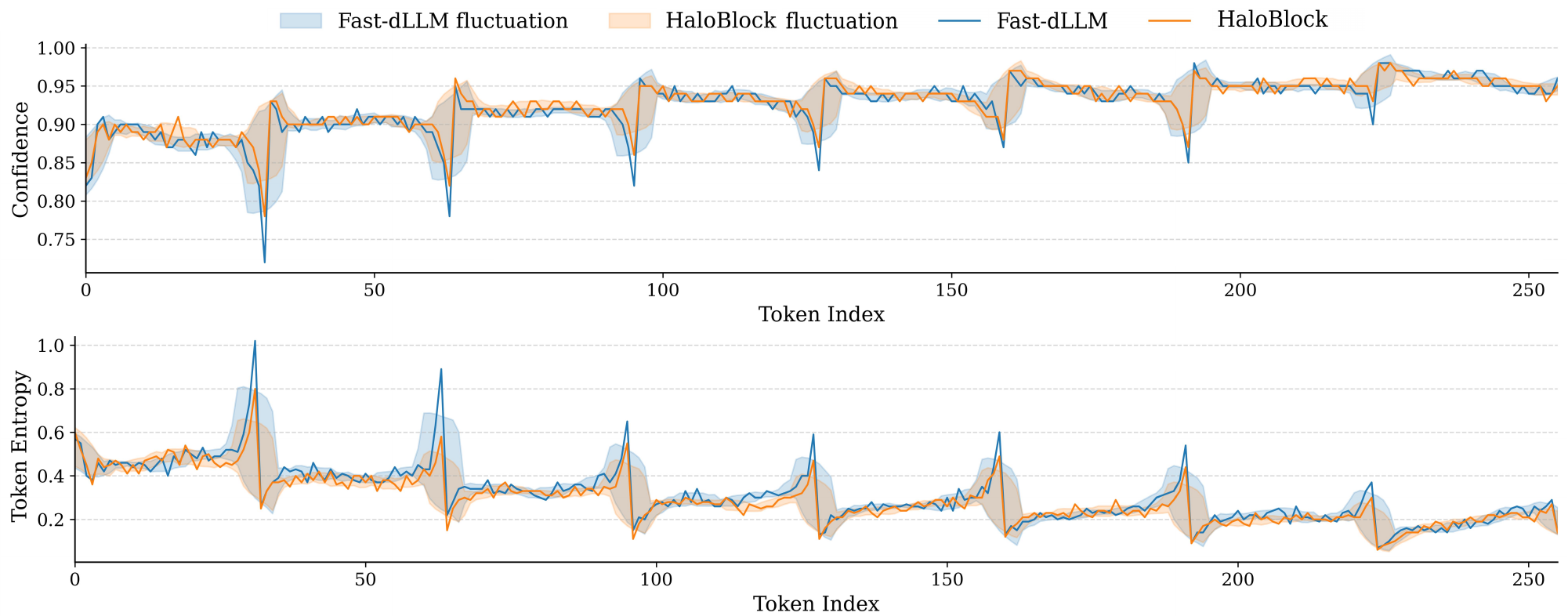}
    \captionsetup{font={small}}
	\caption{Visualization of token confidence and token entropy on LLaDA-8B-Instruct (GSM8K) with generation length $T_x=256$ and block size $t=32$.}
	\label{fig:jump}
 \vspace{0pt}
\end{figure*}

\subsection{Empirical Analysis}
\label{sec: sec1}
\textbf{Block-wise decoding vs. full sequence decoding}.
Block-wise decoding is semi-autoregressive.
It can be seen as an intermediate case between $t=1$ and $t=T_x$, i.e., $1\leq t \leq T_x$.
Generally speaking, the sequence $X$ can be viewed as a preceding subsequence of a longer generating sequence. 
Once $X$ is generated, it can be seen as the prefix condition text to generate the subsequent text.
From this perspective, each decoding block is functionally equivalent to a miniature full-sequence generation task.
Both block-wise decoding and full sequence decoding follow the same discrete diffusion denoising, and they are conditioned on left-side prefix context.
This leads to a conclusion: the limitations observed in full-sequence diffusion decoding will not be eliminated by sequence splitting; instead, they will manifest within each individual block on a smaller scale.


For full-sequence dLLM inference, a well-known intrinsic limitation is the position-dependent disparity in per-token decoding difficulty.
Because of the sequential nature of natural language, tokens near the left side of the sequence benefit from richer accumulated contextual cues from the prefix, which reduces denoising difficulty and leads to faster convergence \citep{AR-Diffusion}.
Tokens closer to the end of the sequence, by contrast, have access to far fewer contextual information and thus require more denoising iterations to reach comparable confidence.
This uneven difficulty distribution, referred to as the left-dominance effect in this work, is a native property of diffusion-based language generation and becomes more pronounced as the sequence length increases.

\textbf{Left-dominance effect}.
In \figref{fig:example}(a), we visualize the per-token Number of Function Evaluations (NFE) when tokens are sampled and unmasked.
It shows that NFE grows linearly as token index increases inside a block.
Although the confidence-aware parallel decoding strategy \citep{llada,fast-dllm} favors the high-confidence token more, which does not introduce position bias, the number of forward passes required for token sampling still exhibits a clear left-dominance effect.
The tokens near the right-side of the block require much more iteration to converge.
This trend holds for all blocks and results in a sawtooth‑shaped variation in per-token NFE across token index.
As a result, it may easily lead to false prediction near the right edge of the block.
An example is illustrated in \figref{fig:example}(b), where the prediction errors occur at a non-trivial frequency in the right area.

\textbf{Boundary contextual discontinuity}. 
The contextual discontinuity is presumably caused by the asynchronous KV cache updates at the block boundary \citep{fast-dllm}.
To better analyze this phenomenon, we visualize the token confidence and token entropy for every position of the sequence and average them over all samples.
As shown in \figref{fig:jump}, there are apparent confidence jumps and entropy jumps at the block boundaries.
Notably, token confidence and entropy are relatively stable for most positions in the block, and only jump sharply near the block end.
This suggests that the linear growth in NFE merely reflects a gradual increase in per-token decoding difficulty, while accumulated denoising errors manifest abruptly at block boundaries.

\begin{figure*}[!t]
	\centering
    \scriptsize
	\setlength{\tabcolsep}{1pt}
	\setlength{\abovecaptionskip}{3pt}
	\begin{overpic}[width=1\textwidth]{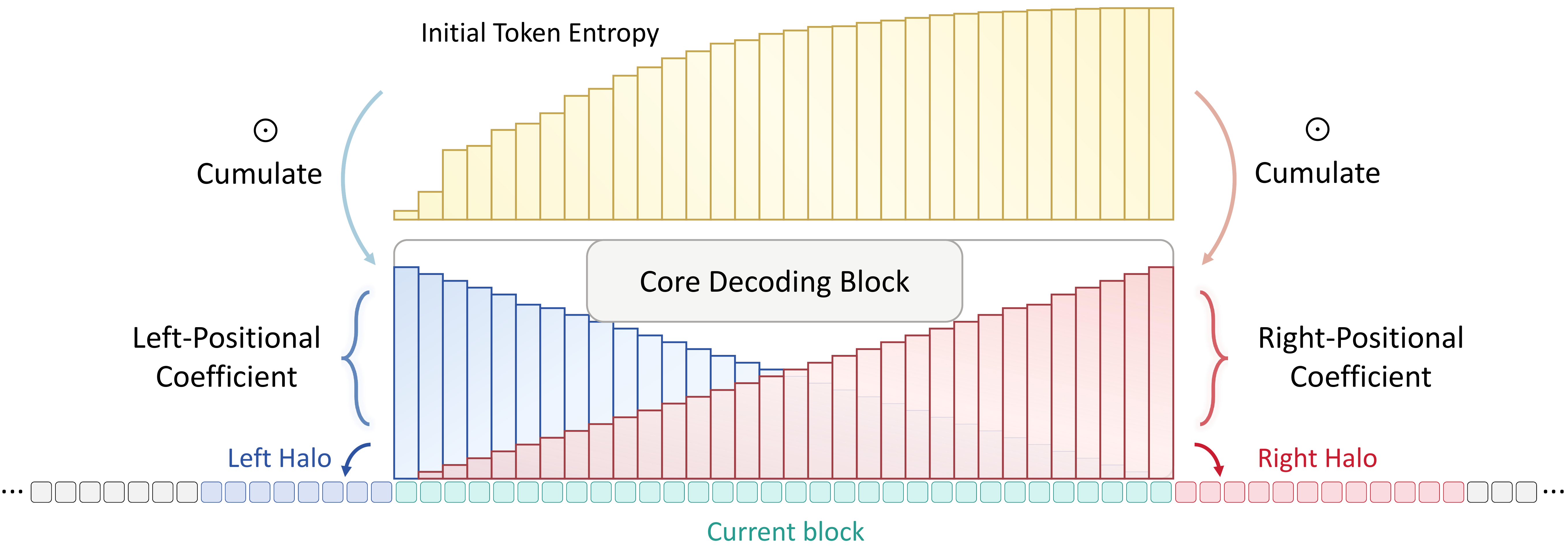}
        \put(25.5,1){$l$}
        \put(73.5,1){$r$}
        \put(12.8,1){$l^{\text{halo}}$}
        \put(92,1){$r^{\text{halo}}$} 
	\end{overpic}
    \captionsetup{font={small}}
	\caption{Computational graph of the HaloBlock. The halo lengths are determined by the bidirectional decoding difficulty, which is cumulated by the positional coefficients and the initial token entropy.}
	\label{fig:haloblock}
 \vspace{0pt}
\end{figure*}

\subsection{HaloBlock}

Motivated by the aforementioned limitations of dLLM decoding, we introduce HaloBlock, a difficulty-aware block-wise decoding framework for dLLM inference.
A halo is an auxiliary token region appended to both sides of the decoding block.
Halo tokens participate in KV-cache updates during iterative denoising, faciliting dynamic contextual cues update and supporting the current block.
Let $X_i=S[l_i:r_i]$ be the $i$-th block with left edge $l_i$ and right edge $r_i$.
The dLLM $\mathcal{M}$ infers the sequence $S$ and outputs the full logits $z$ and the full KV cache $Y$.
We first introduce bidirectional halo regions to deliver contextual support for the block:
\begin{equation}
    l^{\text{halo}}_i, r^{\text{halo}}_i=\texttt{BidirectionalHaloAllocation}(l_i, r_i, z[l_i:r_i]), \quad l^{\text{halo}}_i, r^{\text{halo}}_i \leq T_p + T_x,
\end{equation}
where $z[l_i:r_i]$ represents block logits.
In the following, we delve deep into the function of Bidirectional Halo Allocation (\algref{alg:halo_allocation}).

\begin{algorithm}[t]
\caption{Bidirectional Halo Allocation}
\begin{algorithmic}[1]
\setstretch{1.2}
\Statex \textbf{Inputs: } Block edges $l, r$, block logits $z \in \mathbb{R}^{t \times |\mathcal{V}|}$. Inherited parameters: sequence \(S \in \mathcal{V}^{T_p + T_x}\), temperature $\tau$, halo factor $\alpha$.
\Statex \textbf{Output: } HaloBlock edges $l_{\text{halo}}, r_{\text{halo}}$
\State $p \gets \text{Softmax}(z)$
\State $H \gets -\sum p \log p$ \Comment{Token entropy}
\State $\text{pos} \gets \{0, 1, \dots, t-1\}$ \Comment{Position indices within the block}
\State $c_{\text{left}} \gets 1 - \text{pos} / (t-1)$; \quad $c_{\text{right}} \gets \text{pos} / (t-1)$ \Comment{Linear positional coefficients}
\State $U^{\text{left}} \gets  \alpha\sum \left(H \odot c_{\text{left}}\right)$ \Comment{Left halo length}
\State $U^{\text{right}} \gets  \alpha\sum \left(H \odot c_{\text{right}}\right)$ \Comment{Right halo length}

\State $l^{\text{halo}} \gets \max\left(0,\ l - \lfloor U^{\text{left}} \rfloor\right)$; $r^{\text{halo}} \gets \min\left(T_p + T_x,\ r + \lfloor U^{\text{right}} \rfloor\right)$ 

\State \Return $l^{\text{halo}}, r^{\text{halo}}$
\end{algorithmic}
\label{alg:halo_allocation}
\end{algorithm}

The core of this allocation function is to quantify decoding difficulty in a bidirectional manner, so that halo lengths on each side can be matched to the actual contextual demand of tokens near the corresponding boundary.
We use initial token entropy at the first iterative step as the fundamental measure of per-token decoding difficulty.
Given the block logits $z[l_i:r_i]$, we compute the token-level probability distribution via softmax function, and derive entropy for each position:
\begin{equation}\label{eq:entropy}
    H[l_i:r_i] = -\sum_{v\in|\mathcal{V}|} p[l_i:r_i](v) \log p[l_i:r_i](v); \quad p[l_i:r_i] = \text{Softmax}(z[l_i:r_i]).
\end{equation}
To capture the directional demand for context, we introduce a pair of linear positional coefficients for each token, which is given by:
\begin{equation}
    c_{\text{left}}[l_i:r_i] = 1 - \text{pos} / (t-1); \quad c_{\text{right}}[l_i:r_i] = \text{pos} / (t-1), \quad \text{pos} = \{0, 1, \dots, t-1\}.
\end{equation}
These coefficients assign higher weights to tokens closer to the corresponding block boundary, as tokens near the edge rely more heavily on cross-boundary context for accurate denoising.
We then allocate halo length by cumulating the decoding difficulty for both directions on all $[\mathtt{MASK}]$ tokens:
\begin{equation}\label{eq:halo_length}
    U^{\text{left}}_i = \alpha\sum (H[l_i:r_i] \odot c_{\text{left}}[l_i:r_i]); \quad U^{\text{right}}_i = \alpha\sum (H[l_i:r_i] \odot c_{\text{right}}[l_i:r_i]),
\end{equation}
where $\alpha\geq 0$ is introduced to control the scale of halo regions.
Finally, the edges of HaloBlock can be represented as:
\begin{equation}    
l^{\text{halo}}_i = \max\left(0,\ l - \lfloor U^{\text{left}}_i\rfloor\right); \quad
r^{\text{halo}}_i = \min\left(T_p + T_x,\ r + \lfloor U^{\text{right}}_i\rfloor\right),
\end{equation}
where $\lfloor \cdot \rfloor$ denotes rounding operation.
The resulting HaloBlock is represented by $S[l^{\text{halo}}_i:r^{\text{halo}}_i]$ with left edge $l^{\text{halo}}_i$ and right edge $r^{\text{halo}}_i$.
An overview of HaloBlock computational graph is shown in \figref{fig:haloblock}.

\subsection{Block-wise decoding with HaloBlock}

Integrated into the block-wise decoding pipeline, HaloBlock alleviates the two issues identified in our empirical analysis (see Sec. \ref{sec: sec1}) in a unified manner.

For the intra-block left-dominance effect, the adaptive halo allocation provides asymmetric contextual support: the right side, which suffers from higher decoding difficulty, receives a longer halo and thus more contextual cues to reduce denoising uncertainty.
By delivering more context to harder tokens, the difficulty gap within the block is narrowed.
Since the total number of forward iterations is determined by the hardest tokens, this leads to faster convergence and lower overall NFE (will be evidenced in the experimental section).

For boundary contextual discontinuity, the extended halo region allows tokens near block edges to attend to a wider range of contextual tokens during iterative denoising.
Unlike the standard setting where cross-block KV cache stays static, the halo region updates its KV states along with the current block in each iteration.
This synchronized KV cache update alleviates the sharp confidence jump at block boundaries and improves cross-block semantic coherence (refer to the comparison in \figref{fig:jump}).

Notably, our HaloBlock is plug-and-play while preserving the original block-wise decoding workflow.
The core sampling logic, block scheduling, and KV cache mechanism remain intact, ensuring compatibility with existing dLLM inference optimizations.
The full algorithm is provided in \appnref{sec:haloblock}.

\section{Experiments}

\textbf{Models and benchmarks}.
We evaluate HaloBlock on standard dLLMs: LLaDA-8B-Instruct \citep{llada}, Dream-7B-Instruct \citep{dream7b}, and LLaDA-1.5 \citep{llada-1.5}.
We adopt four widely used benchmarks: GSM8K (5-shot) \citep{GSM8K} and MATH (4-shot) \citep{MATH} for mathematical reasoning, as well as MBPP (3-shot) \citep{MBPP} and HumanEval (0-shot) \citep{humaneval} for code generation.
Generation quality is measured by task accuracy.
Generation efficiency is measured by token per second (TPS) and the total inference time for intuitive comparison.
We also report average NFE to show the number of model forward pass.
All experiments run on four NVIDIA H20 GPUs.

\textbf{Evaluation details}. To provide a fair comparison, we unify the inference pipeline for each model to establish a clean block-wise decoding framework for dLLM inference.
For LLaDA series, we use the representative Fast dLLM framework \citep{fast-dllm}, and we implement HaloBlock on top of Fast dLLM.
For Dream-7B-Instruct, we use the dParallel \citep{dParallel} framework, upon which HaloBlock is implemented.
Since different methods may favor different KV cache strategy, we adopt the one that achieves the best task accuracy, which is fixed in comparison.
HaloBlock introduces one hyperparameter: the halo factor $\alpha$ in \eqref{eq:halo_length} to control the scale of halo regions.
We conduct a grid search on GSM8K dataset and set $\alpha=2$, which is fixed in all the evaluation.
All other dLLM inference settings remain unchanged.

To conduct a comprehensive comparison, we compare HaloBlock against five state‑of‑the‑art dLLM inference approaches, consisting of three training‑based methods: dParallel \citep{dParallel}, D2F \citep{D2F}, and d3LLM \citep{d3llm}, as well as two training‑free methods: Fast dLLM \citep{fast-dllm} and AdaBlock \citep{adablock}.
All these competitors represent the most recent advances in efficient dLLM inference.
We follow the work of our predecessors and use the optimal hyper-parameters of these methods.

\begin{table}[t]
    \setlength{\belowcaptionskip}{0.5mm}
    \caption{\footnotesize Quantitative results of dLLM inference on LLaDA-8B-Instruct. The inference time is  measured in average over three indepentent runs.}
    \setlength{\tabcolsep}{10pt}
    \renewcommand\arraystretch{1.05}
    \centering
    \small
    \begin{tabular}{llccccccccccccccc}
        \toprule
        \textbf{Benchmark} & \textbf{Method} & \textbf{Training-Free} & \textbf{Acc.$\uparrow$} & \textbf{TPS$\uparrow$} & \textbf{NFE$\downarrow$} & \textbf{Time$\downarrow$}\\
        \midrule
        \multirow{7}{*}{\makecell[l]{\textbf{GSM8K} \\ \small (5-shot)}} & LLaDA && 78.5 & 4.0 & 256.0 & 5:19:24 \\
        & dParallel \citep{dParallel} && 76.4 & 27.1 & 32.3 & 0:40:30 \\   
        & D2F \citep{D2F} && 73.9 & 56.0 & 47.4 & \textbf{0:13:25} \\
        & d3LLM \citep{d3llm} && 73.5 & 43.0 & \textbf{25.8} & 0:26:12 \\
        & Fast dLLM \citep{fast-dllm} & \checkmark & 76.0 & 58.5 & 62.8 & 0:22:24 \\
        & AdaBlock \citep{adablock} & \checkmark & 78.6 & 34.5 & 91.2 & 0:40:37 \\
        \rowcolor{green!8}
        & \textbf{HaloBlock (Ours)} & \checkmark & \textbf{79.5} & \textbf{61.4} & 59.5 & 0:20:46 \\
        \midrule
        \multirow{7}{*}{\makecell[l]{\textbf{MATH} \\ \small (4-shot)}} & LLaDA && 33.6 & 5.6 & 256.0 & 15:25:44 \\
        & dParallel \citep{dParallel} && 31.5 & 25.0 & 46.2 & 2:40:10 \\  
        & D2F \citep{D2F} && 29.4 & 56.7 & 84.4 & 1:26:09 \\
        & d3LLM \citep{d3llm} && 30.7 & 54.6 & \textbf{35.4} & \textbf{1:16:07} \\
        & Fast dLLM \citep{fast-dllm} & \checkmark & 32.5 & 53.7 & 92.6 & 1:36:34 \\
        & AdaBlock \citep{adablock} & \checkmark & 32.9 & 34.8 & 110.4 & 2:28:41 \\
        \rowcolor{green!8}
        & \textbf{HaloBlock (Ours)} & \checkmark & \textbf{33.2} & \textbf{57.2} & 88.9 & 1:30:33 \\
        \midrule
        \multirow{7}{*}{\makecell[l]{\textbf{HumanEval} \\ \small (0-shot)}} & LLaDA && 42.1 & 10.9 & 256.0 & 0:15:16  \\
        & dParallel \citep{dParallel} && 39.6 & 60.3 & \textbf{31.6} & \textbf{0:01:45} \\
        & D2F \citep{D2F} && 41.5 & 34.5 & 34.4 & 0:02:26 \\
        & d3LLM \citep{d3llm} && 39.6 & 14.3 & 50.2 & 0:03:17 \\
        & Fast dLLM \citep{fast-dllm} & \checkmark & 42.1 & 55.3 & 87.0 & 0:02:53 \\
        & AdaBlock \citep{adablock} & \checkmark & 39.0 & 42.1 & 94.2 & 0:03:47 \\
        \rowcolor{green!8}
        & \textbf{HaloBlock (Ours)} & \checkmark & \textbf{43.3} & \textbf{60.4} & 84.5 & 0:02:38 \\
        \midrule
        \multirow{7}{*}{\makecell[l]{\textbf{MBPP} \\ \small (3-shot)}} & LLaDA && 40.6 & 3.8 & 256.0 & 1:38:24 \\
        & dParallel \citep{dParallel} && 39.8 & 11.5 & \textbf{23.2} & 0:09:28 \\
        & D2F \citep{D2F} && 36.6 & 31.7 & 31.3 & \textbf{0:04:14} \\
        & d3LLM \citep{d3llm} && 36.0 & 29.3 & 20.3 & 0:04:15 \\
        & Fast dLLM \citep{fast-dllm} & \checkmark & 38.8 & 37.2 & 70.6 & 0:08:33  \\
        & AdaBlock \citep{adablock} & \checkmark & 40.8 & 31.2 & 86.8 & 0:12:14 \\
        \rowcolor{green!8}
        & \textbf{HaloBlock (Ours)} & \checkmark & \textbf{41.2} & \textbf{38.9} & 65.7 & 0:08:15 \\
        \bottomrule
    \end{tabular}
    \vspace{0pt}
    \label{tab:llada}
\end{table}

\begin{table}[t]
    \setlength{\belowcaptionskip}{0.5mm}
    \caption{\footnotesize Quantitative results of dLLM inference on Dream-7B-Instruct. The inference time is  measured in average over three indepentent runs.}
    \setlength{\tabcolsep}{10pt}
    \renewcommand\arraystretch{1.05}
    \centering
    \small
    \begin{tabular}{llccccccccccccccc}
        \toprule
        \textbf{Benchmark} & \textbf{Method} & \textbf{Training-Free} & \textbf{Acc.$\uparrow$} & \textbf{TPS$\uparrow$} & \textbf{NFE$\downarrow$} & \textbf{Time$\downarrow$}\\
        \hline
        \multirow{7}{*}{\makecell[l]{\textbf{GSM8K} \\ \small (5-shot)}} & Dream && 82.9 & 19.2 & 256.0 & 1:13:20 \\
        & dParallel \citep{dParallel} && 81.3 & 138.1 & 41.8 & 0:10:11 \\   
        & D2F \citep{D2F} && 77.5 & 50.9 & 48.7 & 0:12:06 \\
        & d3LLM \citep{d3llm} && 81.7 & 162.0 & \textbf{24.0} & \textbf{0:07:58} \\
        & Fast dLLM \citep{fast-dllm} & \checkmark & 77.8 & 88.1 & 66.1 & 0:14:46 \\
        & AdaBlock \citep{adablock} & \checkmark & 77.8 & 94.1 & 66.2 & 0:14:10 \\
        \rowcolor{green!8}
        & \textbf{HaloBlock (Ours)} & \checkmark & \textbf{82.3} & \textbf{173.1} & 38.0 & 0:08:08 \\
        \midrule
        \multirow{7}{*}{\makecell[l]{\textbf{MATH} \\ \small (4-shot)}} & Dream && 38.3 & 7.9 & 256.0 & 11:16:58 \\
        & dParallel \citep{dParallel} && 37.3 & 110.0 & 66.5 & 0:48:33 \\   
        & D2F \citep{D2F} && 30.9 & 52.8 & 82.8 & 1:14:55 \\
        & d3LLM \citep{d3llm} && 37.1 & 37.3 & \textbf{41.8} & 2:23:14 \\
        & Fast dLLM \citep{fast-dllm} & \checkmark & 37.2 & 48.4 & 128.8 & 1:39:28 \\
        & AdaBlock \citep{adablock} & \checkmark & 38.4 & 63.1 & 103.3 & 1:24:05 \\
        \rowcolor{green!8}
        & \textbf{HaloBlock (Ours)} & \checkmark & \textbf{38.3} & \textbf{122.8} & 63.4 & \textbf{0:43:29} \\
        \midrule
        \multirow{7}{*}{\makecell[l]{\textbf{HumanEval} \\ \small (0-shot)}} & Dream && 55.5 & 12.6 & 256.0 & 0:13:55  \\
        & dParallel \citep{dParallel} && 53.7 & 134.5 & 34.1 & 0:01:12 \\   
        & D2F \citep{D2F} && 51.2 & 28.5 & 107.6 & 0:03:15 \\
        & d3LLM \citep{d3llm} && 56.1 & 68.4 & 35.5 & 0:02:25 \\
        & Fast dLLM \citep{fast-dllm} & \checkmark & 51.2 & 37.2 & 89.6 & 0:02:06 \\
        & AdaBlock \citep{adablock} & \checkmark & 54.9 & 61.5 & 68.7 & 0:02:44 \\
        \rowcolor{green!8}
        & \textbf{HaloBlock (Ours)} & \checkmark & \textbf{56.1} & \textbf{149.0} & \textbf{31.6} & \textbf{0:01:06} \\
        \midrule
        \multirow{7}{*}{\makecell[l]{\textbf{MBPP} \\ \small (3-shot)}} & Dream && 57.6 & 17.1 & 256.0 & 0:30:34 \\
        & dParallel \citep{dParallel} && 53.4 & 187.0 & 27.1 & 0:02:50 \\   
        & D2F \citep{D2F} && 53.2 & 36.3 & 67.7 & 0:05:36 \\
        & d3LLM \citep{d3llm} && 54.8 & 111.9 & 31.8 & 0:04:46 \\
        & Fast dLLM \citep{fast-dllm} & \checkmark & 52.8 & 30.7 & 53.3 & 0:05:09 \\
        & AdaBlock \citep{adablock} & \checkmark & \textbf{55.0} & 101.0 & 51.0 & 0:04:59 \\
        \rowcolor{green!8}
        & \textbf{HaloBlock (Ours)} & \checkmark & 54.6 & \textbf{214.5} & \textbf{27.0} & \textbf{0:02:24} \\
        \bottomrule
    \end{tabular}
    \vspace{0pt}
    \label{tab:dream}
\end{table}

\subsection{Experimental Results}

\myPara{Main results}.
We conduct a series of experiments on different dLLMs and various decoding settings.
The results of LLaDA-8B-Instruct are summarized in \tabref{tab:llada}.
We see that HaloBlock balances speed and accuracy very well across all the four benchmarks, outperforming both training-based and training-free efficient dLLM inference methods.
Specifically, HaloBlock produces the best task accuracy among all the advanced strategies of dLLM inference.
Comparing to the base model Fast dLLM, our method consistently improves the task accuracy by +3.5\%, +0.7\%, +1.2\%, and +2.4\% on GSM8K, MATH, HumanEval, and MBPP benchmarks, respectively.
Notably, HaloBlock even achieves faster inference speed and decoding convergence speed.
For example, compared to Fast dLLM, TPS is improved by 2.9 token/s on GSM8K with 3.3 lower average NFE.
The inference time is notably reduced by 7.3\% from 0:22:24 to 0:20:46.
Similarly, \tabref{tab:dream} reports quantitative results on Dream-7B-Instruct.
In terms of accuracy, HaloBlock remains highly competitive with the vanilla Dream model.
Compared to the baseline dParallel, HaloBlock produces +1.0, +1.0, +2.4, +1.2 accuracy gains on GSM8K, MATH, HumanEval, and MBPP, respectively.
For inference efficiency, compared to existing training-free methods, HaloBlock consistently delivers the highest TPS and fastest inference time across all four benchmarks.
Those training‑based approaches that achieve acceleration by reducing the number of model forward passes; however, they generally suffer from non‑negligible accuracy degradation.
By contrast, HaloBlock yields a more favorable speed‑accuracy trade‑off for dLLM inference.
For the experimental results of LLaDA-1.5, kindly refer to \appnref{sec:llada-1.5}.

\textbf{HaloBlock eases the boundary contextual discontinuity.}
To verify how HaloBlock takes effect on block-wise decoding, we visualize per-token denoising confidence and token entropy over a 256-token generation sequence.
As shown in \figref{fig:jump}, we observe consistent and abrupt discontinuities at every block boundary: denoising confidence drops sharply and token entropy spikes upward, as tokens adjacent to the partition lose direct connection to the new contextual information from the neighboring block.
These jumps mark positions where decoding uncertainty rises abruptly, which corresponds to where most generation errors occur in block-wise dLLM.
HaloBlock effectively smooths these boundary-induced fluctuations.
With adaptive halo regions extending the attention window across partitions, the magnitude of confidence drops and entropy spikes at boundaries is substantially reduced, yielding a more stable and continuous generation across the entire generated sequence.

\begin{figure*}[!t]
	\centering
    \scriptsize
	\setlength{\tabcolsep}{1pt}
	\setlength{\abovecaptionskip}{3pt}
	\includegraphics[width=1\textwidth]{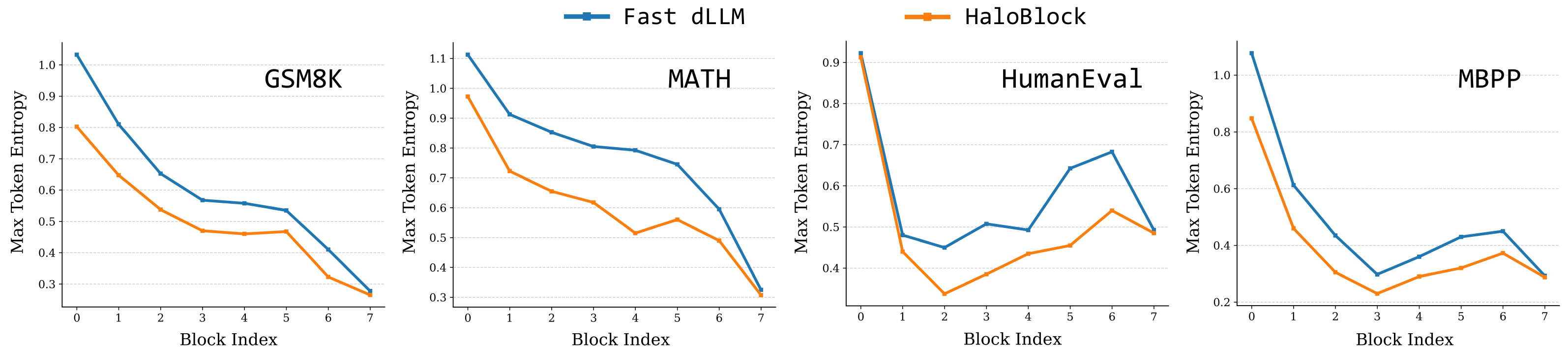}
    \captionsetup{font={small}}
	\caption{Visualization of the max token entropy for each block when the tokens are unmasked. HaloBlock reduces the max token entropy, which indicates lower uncertainty.}
	\label{fig:max-token-entropy}
 \vspace{0pt}
\end{figure*}

\textbf{HaloBlock accelerates decoding convergence by reducing decoding difficulty}.
One might expect that attaching the effective context window via halo regions would incur higher computational overhead per forward pass.
To further understand the source of HaloBlock's efficiency gains, we visualize the maximum token entropy within each block across 8 consecutive blocks, evaluated on four benchmarks: GSM8K, MATH, HumanEval, and MBPP.
As shown in \figref{fig:max-token-entropy}, across all four datasets, HaloBlock consistently yields lower maximum token entropy for every block compared to the Fast dLLM baseline.
In block-wise decoding, the convergence speed of each block is highly correlated with the token of the highest uncertainty, as this token acts as the rate-limiting step for the entire block.
The reduction in maximum per-block token entropy indicates that HaloBlock effectively lowers the decoding difficulty of the hardest tokens within each block, which directly leads to faster convergence.
As a result, we observe that HaloBlock reaches acceptable denoising quality with fewer model forward passes compared to the baseline, which explains why HaloBlock achieves higher throughput despite operating with a slightly larger effective attention scope: it trades a modest increase in per-step compute for a substantial reduction in the number of iterative steps.
More analyses of HaloBlock can be found in \appnref{sec:analysis}.

\begin{table}[t]
    \setlength{\belowcaptionskip}{0.5mm}
    \caption{\footnotesize Quantitative results of dLLM inference with various generation length $T_x$ and block size $t$. The inference time is  measured in average over three indepentent runs.}
    \setlength{\tabcolsep}{6pt}
    \renewcommand\arraystretch{1.05}
    \centering
    \small
    \begin{tabular}{lcccccccccccccccc}
        \toprule
        \multirow{2}{*}{\textbf{Method}} & \multirow{2}{*}{$T_x$} && \multicolumn{3}{c}{$t=8$} && \multicolumn{3}{c}{$t=16$} && \multicolumn{3}{c}{$t=32$} \\
        \cmidrule(lr){4-6}\cmidrule(lr){8-10}\cmidrule(lr){12-14}
        &&& \textbf{Acc.$\uparrow$} & \textbf{TPS$\uparrow$} & \textbf{Time$\downarrow$} && \textbf{Acc.$\uparrow$} & \textbf{TPS$\uparrow$} & \textbf{Time$\downarrow$} && \textbf{Acc.$\uparrow$} & \textbf{TPS$\uparrow$} & \textbf{Time$\downarrow$}\\
        \midrule
        LLaDA & 256 && 80.5 & 4.4 & 4:49:38 && 79.2 & 4.5 & 4:48:55 && 78.5 & 4.0 & 5:19:24 \\
        Fast dLLM & 256 && 78.9 & 24.1 & 0:52:40 && 77.4 & 35.9 & 0:35:32 && 76.0 & 58.5 & 0:22:24\\
        HaloBlock & 256 && 80.7 & 25.5 & 0:50:11 && 79.8 & 38.7 & 0:33:00 && 79.5 & 61.4 & 0:20:46\\
        \midrule
        LLaDA & 512 && 80.4 & 2.1 & 11:30:26 && 77.7 & 2.1 & 11:31:15 && 76.1 & 2.5 & 10:14:10 \\
        Fast dLLM & 512 && 76.7 & 15.5 & 1:36:27 && 76.9 & 25.5 & 0:56:01 && 74.5 & 40.4 & 0:37:17 \\
        HaloBlock & 512 && 78.8 & 15.6 & 1:35:38 && 79.5 & 27.1 & 0:54:35 && 77.5 & 42.7 & 0:34:53 \\
        \bottomrule
    \end{tabular}
    \vspace{0pt}
    \label{tab:ablation_length}
\end{table}

\subsection{Ablation Study}

\textbf{Performance across different generation length}.
We experiment HaloBlock under two widely adopted generation lengths, $256$ and $512$, to verify its robustness across output length.
As shown in \tabref{tab:ablation_length}, doubling the sequence length reduces throughput for all three methods, as each forward pass operates on a longer sequence and incurs higher computational cost.
Across generation length 256 and 512, HaloBlock boosts the generation quality while achieving inference speedup.
Meanwhile, the throughput advantage of HaloBlock over Fast dLLM remains consistent across lengths, indicating that the overhead of halo allocation is negligible and does not scale with total sequence length.

\textbf{Performance across different block size}.
We further examine how decoding performance varies with block size $t\in\{8, 16, 32\}$.
It can be seen from \tabref{tab:ablation_length} that larger block sizes yield higher throughput, since more tokens are denoised in parallel per forward evaluation.
However, larger blocks also cause accuracy degradation: when $t$ increases from 8 to 32, the task accuracy of Fast dLLM drops 2.9 at $T_x=256$ and 2.2 at $T_x=512$.
HaloBlock effectively improves this accuracy-speed balance.
Across all tested block sizes, it recovers the majority of the accuracy drops, while retaining or even improving inference efficiency.

\textbf{Effect of halo factor $\alpha$.}
In \eqref{tab:halo-factor}, HaloBlock introduces one hyper-parameter $\alpha$ to control the scale of halo regions.
We investigate the effect of HaloBlock under various halo factors.
\tabref{tab:halo-factor} reports the experimental results on LLaDA-8B-Instruct and Dream-7B-Instruct.
We can see that all the choices of $\alpha$ improve the performance of dLLM.
In practice, we empirically set $\alpha=2$ for all models.

\begin{wrapfigure}{r}{0.5\textwidth}
  \vspace{-26pt}
  \begin{center}
		\setlength{\fboxrule}{0pt}
		\fbox{\includegraphics[width=0.5\textwidth]{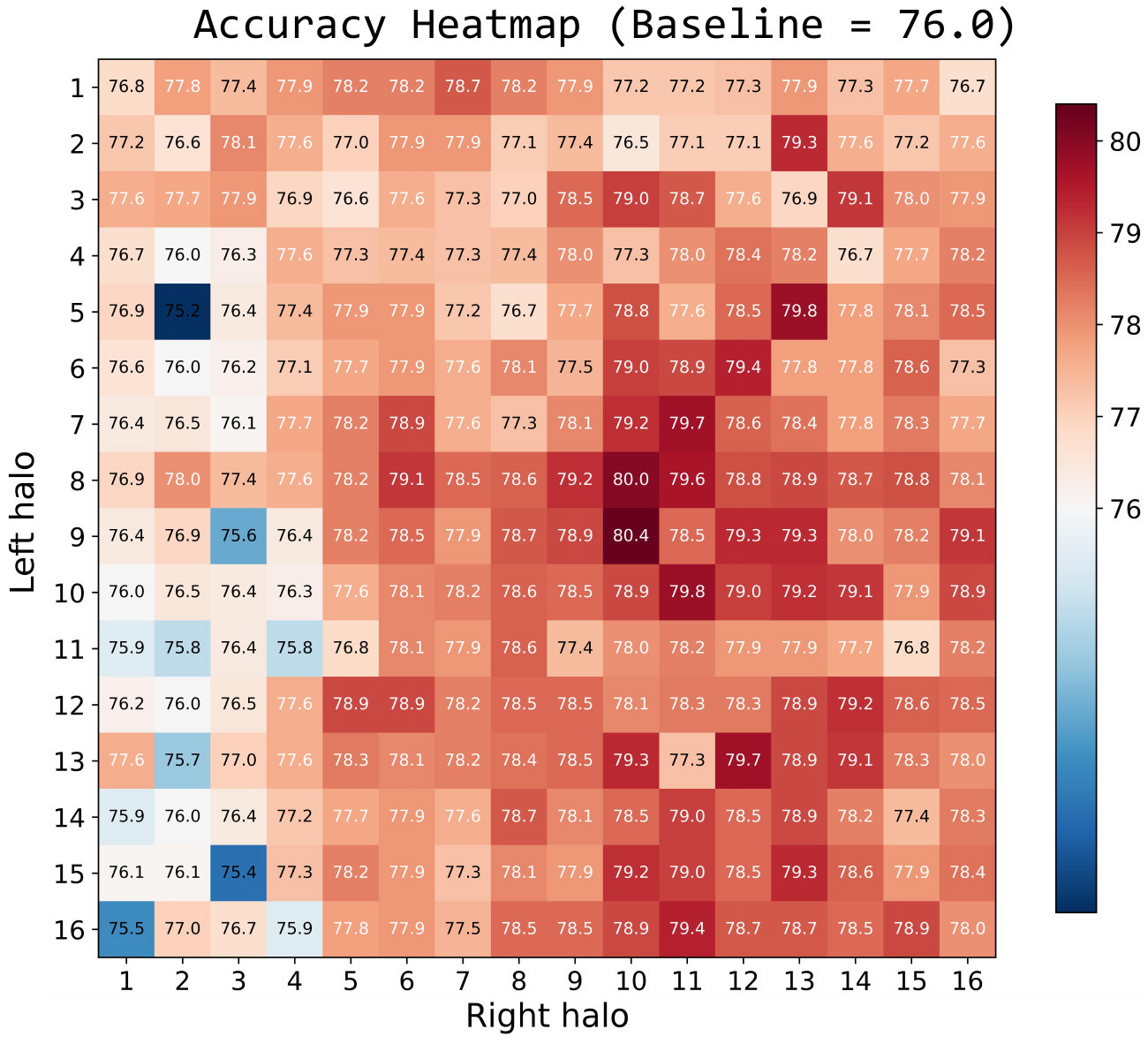}}
	\end{center}
	\captionsetup{font={small}}
      \vspace{-22pt}
	\caption{Visualization of accuracy heatmap with static halo region on LLaDA-8B-Instruct (GSM8K).}
  \label{fig:heatmaps}
  \vspace{-8pt}
\end{wrapfigure}

\textbf{Halo region: difficulty-aware vs. static}.
We perform a grid search over static halo configurations, where left and right halo lengths are varied respectively from 1 to 16 tokens.
As shown in \figref{fig:heatmaps}, static halos improve over the 76.0 baseline across most settings, with peak accuracy at around $(9, 10)$.
Our HaloBlock reaches 79.5\% accuracy without the need of laboriously tuning the left and the right halo length.
An intriguing finding is the strong asymmetry between left and right halo sensitivity.
A short right halo ($U^{\text{right}}\leq5$) leads to marked accuracy drops, and even worse performance than having no halo at all in some configurations.
This asymmetry presumably stems from the directional nature of sequence generation.
The left boundary of a block adjoins already decoded tokens with fully available context, thus it is less dependent on halo extension.
The right boundary faces mask tokens, where native contextual information is inherently scarce, making adequate right halo coverage critical for stable decoding.
Our difficulty-aware approach adaptively prioritizes high-difficulty positions and allocates halo capacity efficiently.

\textbf{Compatibility of HaloBlock}.
We then verify that HaloBlock is orthogonal to some recent dLLM inference optimizations.
For example, adaptive block size methods such as AdaBlock \citep{adablock}, dynamically assigns block sizes via semantic delimiters.
Due to its better alignment between block size and semantic step, AdaBlock can effectively enhance the sampling quality.
We then adapt HaloBlock to AdaBlock. The results in \tabref{tab:compatibility} show that further improvements can be achieved in both generation quality and inference efficiency.
These results show the consistent effectiveness of HaloBlock: better generation accuracy, faster decoding convergence, and lower inference cost.

\begin{table}[!tb]
\setlength{\belowcaptionskip}{0.5mm}
\begin{minipage}[t]{0.52\linewidth}
\caption{Ablation study on halo factor $\alpha$. The results are reported by task accuracy and inference time.}
\label{tab:halo-factor}
\renewcommand\arraystretch{1.05}
\setlength{\tabcolsep}{4pt}
\centering
    \small
    \begin{tabular}{lcccccccccccccccc}
        \toprule
        Model & Baseline & $\alpha=1$ & $\alpha=2$ & $\alpha=3$ & $\alpha=4$ \\
        \midrule
        \multirow{2}{*}{LLaDA} & 76.0 & 78.0 & 79.5 & 79.2 & 78.2 \\
        & 0:22:24 & 0:20:32 & 0:20:46 & 0:21:22 & 0:22:17 \\
        \midrule
        \multirow{2}{*}{Dream} & 81.3 & 81.8 & 82.3 & 82.0 & 81.8 \\
        & 0:10:11 & 0:07:49 & 0:08:08 & 0:08:44 & 0:08:48 \\
        \bottomrule
    \end{tabular}
\end{minipage}
\,\,
\begin{minipage}[t]{0.46\linewidth}
\caption{Performance of HaloBlock upon more block-wise decoding methods.}
\label{tab:compatibility}
\renewcommand\arraystretch{1.05}
\setlength{\tabcolsep}{4pt}
\centering
    \small
    \begin{tabular}{lcccccccccccccccc}
        \toprule
        Model & Acc.$\uparrow$ & TPS$\uparrow$ & NFE$\downarrow$ & Time$\downarrow$ \\
        \midrule
        AdaBlock & 78.6 & 34.5 & 91.2 & 0:40:37 \\
        AdaBlock + Ours & 79.4 & 35.6 & 87.5 & 0:39:33 \\
        \midrule
        dParallel & 81.3 & 138.1 & 41.8 & 0:10:11 \\
        dParallel + Ours & 82.3 & 173.1 & 38.0 & 0:08:08 \\
        \bottomrule
    \end{tabular}
\end{minipage}
\vspace{0pt}
\end{table}

\section{Conclusion}

In this paper, we presented HaloBlock, a simple yet effective method to improve the inference speed and accuracy of dLLMs by alleviating the issues of left-dominance effect and boundary contextual discontinuity.
HaloBlock quantifies token decoding difficulty via bidirectional cumulative entropy and use it to adaptively appending halo regions to both sides of each block.
Extensive evaluations on LLaDA and Dream models demonstrated that HaloBlock achieves consistent generation quality improvement, as well as faster decoding convergence and lower inference time cost.
This work has implications for how we model token decoding difficulty and design contextual support algorithms in block-wise diffusion large language model inference.

\textbf{Limitations}.
While simple and effective, as many other dLLM acceleration methods, HaloBlock still relies on block-wise decoding and KV cache.
%
In addition, currently we apply HaloBlock to the popular LLaDA, LLaDA-1.5, and Dream models.
In the future, we will adapt HaloBlock to the recently developed dLLMs such as LLaDA-2.0 \citep{llada2}, SDAR \citep{SDAR}, d3LLM \citep{d3llm} and DMPO \citep{DMPO}, etc.


{\small
\bibliography{ref}
}

\appendix

\clearpage
\setcounter{table}{0}
\setcounter{equation}{0}
\setcounter{figure}{0}
\renewcommand{\thetable}{\thesection.\arabic{table}}
\renewcommand{\theequation}{\thesection.\arabic{equation}}
\renewcommand{\thefigure}{\thesection.\arabic{figure}}
\begin{center}
    {\LARGE\bfseries Appendix}
\end{center}

\section{Block-wise Decoding with HaloBlock}\label{sec:haloblock}
We present the formal algorithm for the proposed HaloBlock in \algref{alg:haloblock}, where we adopt Fast dLLM \citep{fast-dllm} as an example baseline framework.

\begin{algorithm}[!tb]
\caption{HaloBlock}
\begin{algorithmic}[1]
\setstretch{1.2}
\Statex \textbf{Inputs: } Model \(\mathcal{M}\), prompt sequence \(P \in \mathcal{V}^{T_p}\), refinement steps $s$, generation length \(T_x\), block size \(t\), the number of blocks $b$, temperature $\tau$, halo factor $\alpha$.
\Statex \textbf{Output: } Generated full sequence \(S = [P, X]\in \mathcal{V}^{T_p + T_x}\)
\State \(S \gets [P, \underbrace{[\mathtt{MASK}],\cdots, [\mathtt{MASK}]}_{T_x}]\) \Comment{Initialize sequence}
\For{\(i = 1\) \textbf{to} \(b\)}
\State \(l \gets T_p + (i-1) \cdot t\); \quad \(r \gets l + t\) \Comment{Left and right edges of block \(X_i\)}
\State \(\mathcal{I} \gets (S[l:r] = [\mathtt{MASK}])\)
\State \(\{n_k\}_{k=1}^{s} \gets \texttt{AllocQuota}(\mathcal{I}, s)\) \Comment{Allocate unmasking quota per step}
\State \(\mathcal{R} \gets \mathbf{1}^{T_p + T_x}\) \Comment{KV cache update indicator}
\State \(z, Y \gets \mathcal{M}(S, \mathcal{R})\) \Comment{Obtain full logits and full KV cache}
\State \(S \gets \texttt{Sampling}(z[l:r], \mathcal{I}, S, n_1)\) \Comment{Decode and update \(X\)}
\State \(l^{\text{halo}}, r^{\text{halo}} = \texttt{BidirectionalHaloAllocation}(l, r, z[l:r])\)
\State \(\mathcal{R} \gets \mathbf{0}^{T_p + T_x}\); \quad \(\mathcal{R}[l^{\text{halo}} : r^{\text{halo}}] \gets \mathbf{1}\) \Comment{Identify KV cache update indicator}
\For{\(j = 2\) \textbf{to} \(s\)}
\State \(\mathcal{I} \gets (S[l:r] = [\mathtt{MASK}])\)
\If{\(\sum\mathcal{I} = 0\)}
\State \textbf{break} \Comment{Early exit when block is fully unmasked}
\EndIf
\State \(z_j, Y \gets \mathcal{M}(S[l^{\text{halo}}:r^{\text{halo}}], Y, \mathcal{R})\) \Comment{Block-level logits with updated KV cache}
\State \(S \gets \texttt{Sampling}(z_j[l-l^{\text{halo}}:r-l^{\text{halo}}], \mathcal{I}, S, n_j)\) \Comment{Decode and update \(X\)}
\EndFor\EndFor\State \Return \(S\)
\end{algorithmic}
\label{alg:haloblock}
\end{algorithm}

\section{Experimental Results on LLaDA-1.5}\label{sec:llada-1.5}

\begin{table}[h]
    \setlength{\belowcaptionskip}{0.5mm}
    \caption{\footnotesize Quantitative results of dLLM inference on LLaDA-1.5. The inference time is  measured in average over three indepentent runs.}
    \setlength{\tabcolsep}{10pt}
    \renewcommand\arraystretch{1.1}
    \centering
    \small
    \begin{tabular}{llccccccccccccccc}
        \toprule
        \textbf{Benchmark} & \textbf{Method} & \textbf{Training-Free} & \textbf{Acc.$\uparrow$} & \textbf{TPS$\uparrow$} & \textbf{NFE$\downarrow$} & \textbf{Time$\downarrow$}\\
        \midrule
        \multirow{4}{*}{\makecell[l]{\textbf{GSM8K} \\ \small (5-shot)}} & LLaDA-1.5 \citep{llada-1.5} && 80.5 & 3.8 & 256.0 & 5:14:26  \\
        & Fast dLLM \citep{fast-dllm} & \checkmark & 80.3 & 46.3 & 84.0 & 0:25:15 \\
        & AdaBlock \citep{adablock} & \checkmark & 79.8 & 31.7 & 88.1 & 0:37:06 \\
        \rowcolor{green!8}
        & \textbf{HaloBlock (Ours)} & \checkmark & \textbf{81.1} & \textbf{49.4} & \textbf{79.4} & \textbf{0:24:03} \\
        \midrule
        \multirow{4}{*}{\makecell[l]{\textbf{MATH} \\ \small (4-shot)}} & LLaDA-1.5 \citep{llada-1.5} && 33.2 & 5.2 & 256.0 & 13:12:47 \\
        & Fast dLLM \citep{fast-dllm} & \checkmark & 32.5 & 40.6 & 106.5 & 1:51:09 \\
        & AdaBlock \citep{adablock} & \checkmark & 33.0 & 42.9 & 108.1 & 2:04:13 \\
        \rowcolor{green!8}
        & \textbf{HaloBlock (Ours)} & \checkmark & \textbf{33.3} & \textbf{45.9} & \textbf{100.3} & \textbf{1:38:10} \\
        \midrule
        \multirow{4}{*}{\makecell[l]{\textbf{HumanEval} \\ \small (0-shot)}} & LLaDA-1.5 \citep{llada-1.5} && 43.3 & 4.4 & 256.0 & 0:12:23  \\
        & Fast dLLM \citep{fast-dllm} & \checkmark & 35.4 & 15.1 & 106.2 & 0:03:57 \\
        & AdaBlock \citep{adablock} & \checkmark & 37.8 & 12.3 & 123.5 & 0:04:05 \\
        \rowcolor{green!8}
        & \textbf{HaloBlock (Ours)} & \checkmark & \textbf{39.0} & \textbf{18.0} & \textbf{101.3} & \textbf{0:03:13} \\
        \midrule
        \multirow{4}{*}{\makecell[l]{\textbf{MBPP} \\ \small (3-shot)}} & LLaDA-1.5 \citep{llada-1.5} && 39.2 & 1.4 & 256.0 & 1:27:34 \\
        & Fast dLLM \citep{fast-dllm} & \checkmark & 35.2 & 20.2 & 45.3 & 0:06:06  \\
        & AdaBlock \citep{adablock} & \checkmark & \textbf{37.6} & 14.1 & 52.0 & 0:09:04 \\
        \rowcolor{green!8}
        & \textbf{HaloBlock (Ours)} & \checkmark & \textbf{37.6} & \textbf{25.4} & \textbf{42.7} & \textbf{0:05:25} \\
        \bottomrule
    \end{tabular}
    \vspace{0pt}
    \label{tab:llada-1.5}
\end{table}

We compare HaloBlock with the training-free dLLM inference methods, Fast dLLM \citep{fast-dllm} and AdaBlock \citep{adablock} on LLaDA-1.5.
The quantitative results are presented in \tabref{tab:llada-1.5}.
Compared to Fast dLLM and AdaBlock,  HaloBlock yields accuracy gains and better efficiency under all the four benchmarks.
This again demonstrates the effectiveness of our proposed method.

\section{More Analysis on HaloBlock}\label{sec:analysis}

\textbf{HaloBlock eases the decoding difficulty.}
To further analyze the behavior of using HaloBlock on dLLM inference, we introduce cumulative decoding uncertainty (CDU) to represent the decoding difficulty for the sequence $S[l:r]$,
\begin{equation}
    \text{CDU}[l:r] = \sum_{j=1}^s||H_j[l:r]\cdot\mathcal{I}_j||_2,\quad j\in\{1,2,\cdots,s\},
\end{equation}
where $\mathcal{I}_j=(S_j[l:r]=[\mathtt{MASK}])$ denotes the iterative indicator of the sequence $S[l:r]$ checking whether the tokens are unmasked.
Therefore, CDU calculates the cumulative token entropy from the beginning to the final unmasking.
For easy tokens, the unmasking is quickly achieved, resulting in a low CDU, while for hard tokens, CDU is large as they need multiple iteration to converge.

\begin{figure*}[!t]
    \small
	\centering
	\setlength{\tabcolsep}{1pt}
	\setlength{\abovecaptionskip}{3pt}
	\begin{overpic}[width=1\textwidth]{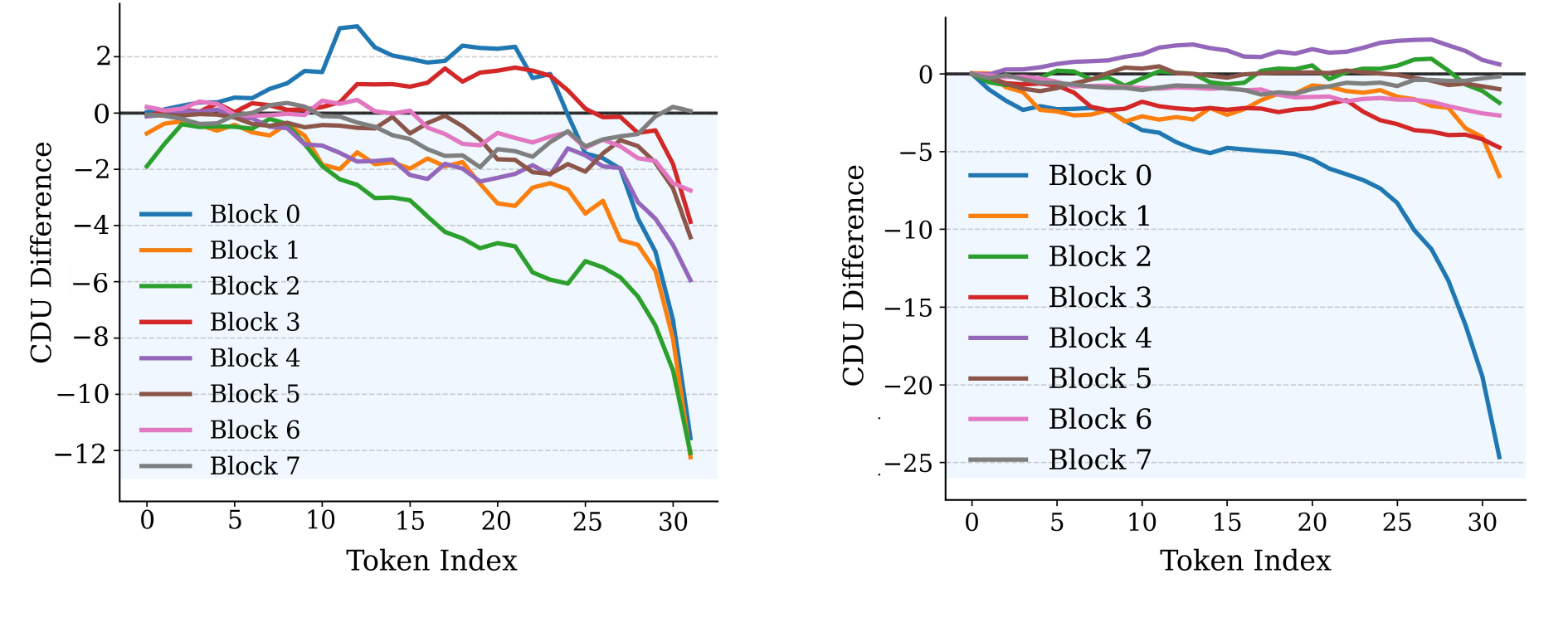}
        \put(23,1){(a) LLaDA}
        \put(74.5,1){(b) Dream}
	\end{overpic}
	\caption{Visualization of CDU difference for every block. The difference is calculated by subtracting the CDU of our HaloBlock from the CDU of Fast dLLM. The block size $t=32$.}
	\label{fig:CDU}
 \vspace{0pt}
\end{figure*}

To showcase how our method affects decoding difficulty, we visualize the CDU difference between our HaloBlock and Fast dLLM.
We calculate the CDU difference by subtracting the CDU of our HaloBlock from the CDU of Fast dLLM.
As shown in \figref{fig:CDU}, when the token approaches the right edge of the block, our method can generally produce lower CDU as the CDU difference decreases to negatives.
This indicates that our method can effectively reduce the decoding difficulty of hard tokens located on the right side of the block.
As a result, faster decoding convergence is achieved.
The average NFE is reduced as \tabref{tab:llada}, \tabref{tab:dream}, and \tabref{tab:llada-1.5} show.

\begin{figure*}[!t]
    \small
	\centering
	\setlength{\tabcolsep}{1pt}
	\setlength{\abovecaptionskip}{3pt}
	\begin{overpic}[width=0.94\textwidth]{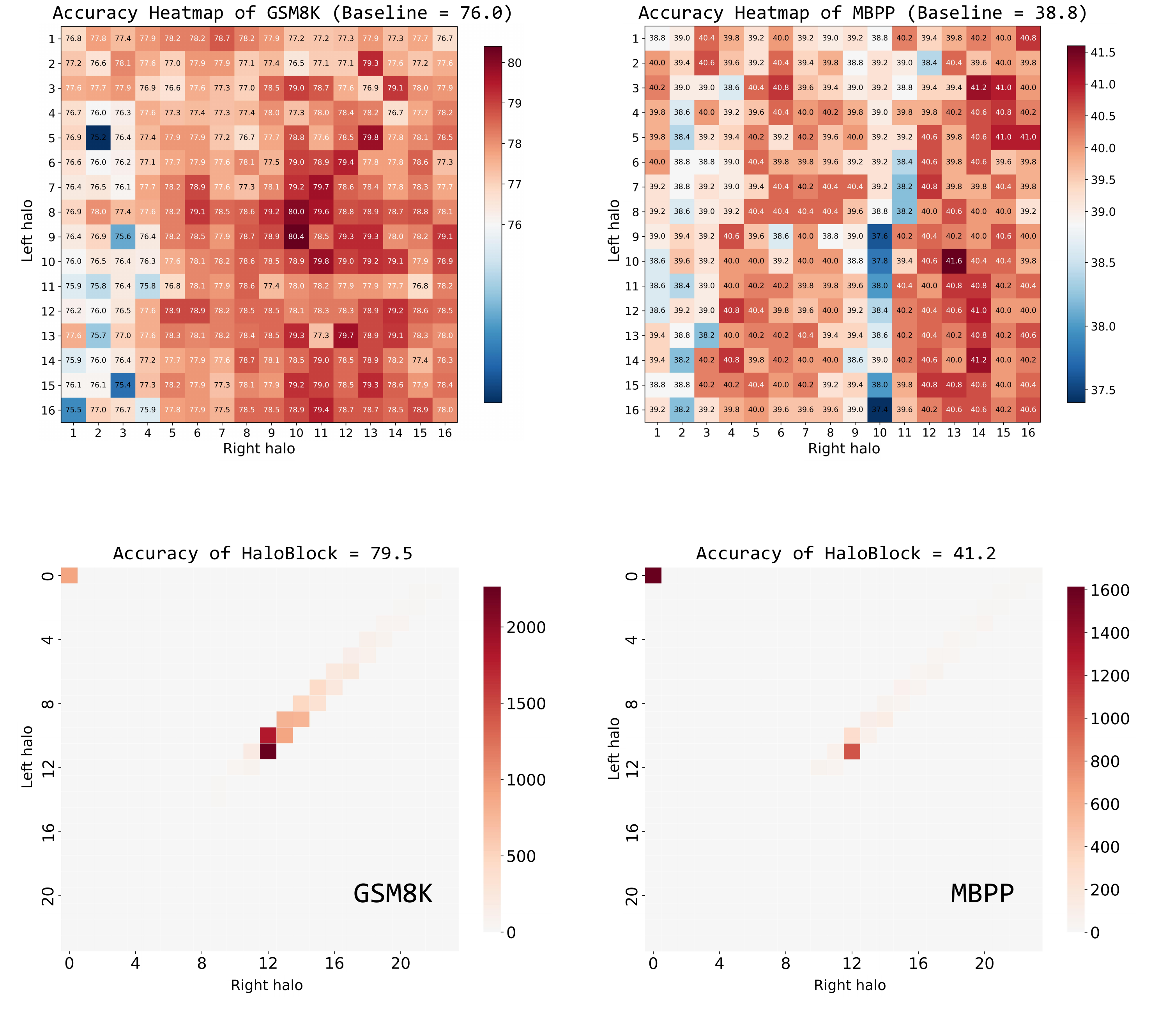}
        \put(21.3,47.9){(a)}
        \put(72,47.8){(b)}
        \put(22,1){(c)}
        \put(72.6,1){(d)}
	\end{overpic}
	\caption{Visualization of accuracy heatmap with static halo region and the statistical count of halo length by our HaloBlock. Our adaptive halo allocation generally performs better than static halos.}
	\label{fig:heatmaps-gsm8k-mbpp}
 \vspace{0pt}
\end{figure*}

\textbf{HaloBlock is hyper-parameter free.}
In \figref{fig:heatmaps} of the main paper, we plot the accuracy heatmap with static halo region on GSM8K dataset.
We here supplement the accuracy heatmap on MBPP \citep{MBPP}.
Comparing the two accuracy heatmaps in \figref{fig:heatmaps-gsm8k-mbpp}(a) and \figref{fig:heatmaps-gsm8k-mbpp}(b), one can see that the optimal halo configuration on GSM8K may not be the optimum on MBPP.
Particularly, the accuracy peak of GSM8K is at $(9,10)$, which intriguingly turns out to be a performance degradation point on MBPP, yielding an accuracy of 37.6\% even lower than the baseline 38.8\%.
This observation demonstrates that static fixed‑size halo configurations cannot generalize across different tasks, as optimal halo lengths are task‑dependent and cannot be manually pre‑tuned for all downstream scenarios.
By comparison, our adaptive halo allocation dynamically determines left‑ and right‑halo lengths based on bidirectional token decoding difficulty.
It obviates dataset‑specific hyperparameter tuning and achieves robust performance across different reasoning and code‑generation benchmarks.

\textbf{Statistical count of halo length}.
We plot the statistical count of halo length on GSM8K and MBPP in \figref{fig:heatmaps-gsm8k-mbpp}(c) and (d).
It shows that our HaloBlock produces dynamic halo length.
For easy tokens, the halo lengths are mostly fall into 0, because the cumulative entropies approach 0.
For difficult tokens, the model’s output probability distribution is more dispersed, leading to higher entropy.
HaloBlock therefore allocates larger halo length for these hard tokens, which account for a non-trivial proportion of all tokens.
In addition, an asymmetric distribution of halo allocation can also be observed: the right halo length is generally longer than the left counterpart, due to the left-dominance effect.

\begin{figure*}[!t]
    \small
	\centering
	\setlength{\tabcolsep}{1pt}
	\setlength{\abovecaptionskip}{3pt}
    \begin{minipage}{\linewidth}
	\begin{overpic}[width=1\textwidth]{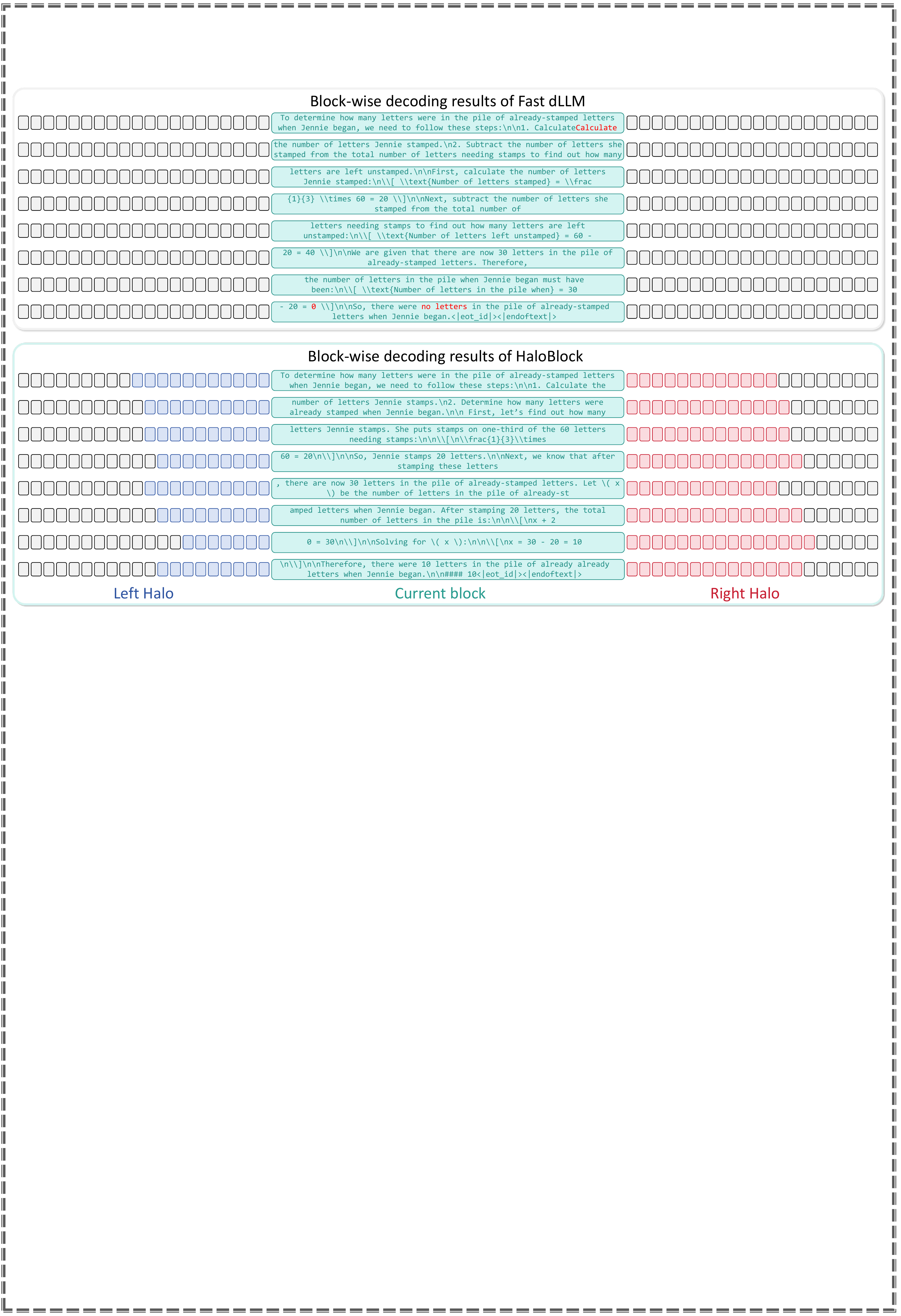}

        \put(1,98){\parbox[t]{0.97\linewidth}{
        Prompt: Jennie is helping at her mom's office. She has a pile of 60 letters needing stamps, and a pile of letters already stamped. She puts stamps on one-third of the letters needing stamps. If there are now 30 letters in the pile of already-stamped letters, how many were in that pile when Jennie began?}}
            
        \put(3,52.5){\footnotesize Note: In the above, the reasoning errors are highlighted in \textcolor{red}{red} and the missing predictions are highlighted in \textbf{bold black}.}

        \put(1,50){\parbox[t]{0.47\linewidth}{
        \textbf{Answer by Fast dLLM}: To determine how many letters were in the pile of already-stamped letters when Jennie began, we need to follow these steps:\\
        
        1. Calculate the number of letters Jennie stamped.
        
        2. Subtract the number of letters she stamped from the total number of letters needing stamps to find out how many letters are left unstamped.\\
        
        First, calculate the number of letters Jennie stamped:
        \[ \text{Number of letters stamped} = \frac{1}{3} \times 60 = 20 \]
        
        Next, subtract the number of letters she stamped from the total number of letters needing stamps to find out how many letters are left unstamped:
        \[ \text{Number of letters left unstamped} = 60 - 20 = 40 \]
        We are given that there are now 30 letters in the pile of already-stamped letters. Therefore, the number of letters in the pile when Jennie began must have been:
        
        \[ \text{Number of letters in the pile when} = \textcolor{red}{30 - 20 = 0} \]
        So, there were \textcolor{red}{no letters} in the pile of already-stamped letters when Jennie began.
        }}

        \put(34.5,50){\parbox[t]{0.47\linewidth}{
        \textbf{Answer by HaloBlock}: To determine how many letters were in the pile of already-stamped letters when Jennie began, we need to follow these steps:\\
        
        1. Calculate the number of letters Jennie stamps.
        
        2. Determine how many letters were already stamped when Jennie began.\\
        
        First, let’s find out how many letters Jennie stamps. She puts stamps on one-third of the 60 letters needing stamps:
        \[
        \frac{1}{3}\times 60 = 20
        \]
        
        So, Jennie stamps 20 letters.\\
        
        Next, we know that after stamping these letters, there are now 30 letters in the pile of already-stamped letters. Let \( x \) be the number of letters in the pile of already-stamped letters when Jennie began. After stamping 20 letters, the total number of letters in the pile is:
        
        \[
        x + 20 = 30
        \]
        
        Solving for \( x \):
        
        \[
        x = 30 - 20 = 10
        \]
        
        Therefore, there were 10 letters in the pile of already-stamped letters when Jennie began.
        
        \#\#\#\# 10
        }}
	\end{overpic}
    \end{minipage}
    \\[2pt]
	\caption{\textbf{Example 1}: Comparison of inference results between the baseline Fast dLLM and HaloBlock. We visualize the dynamic halo length in colors. Reasoning errors are highlighted in \textcolor{red}{red}.}
	\label{fig:example1}
 \vspace{0pt}
\end{figure*}

\begin{figure*}[!t]
    \small
	\centering
	\setlength{\tabcolsep}{1pt}
	\setlength{\abovecaptionskip}{3pt}
    \begin{minipage}{\linewidth}
	\begin{overpic}[width=1\textwidth]{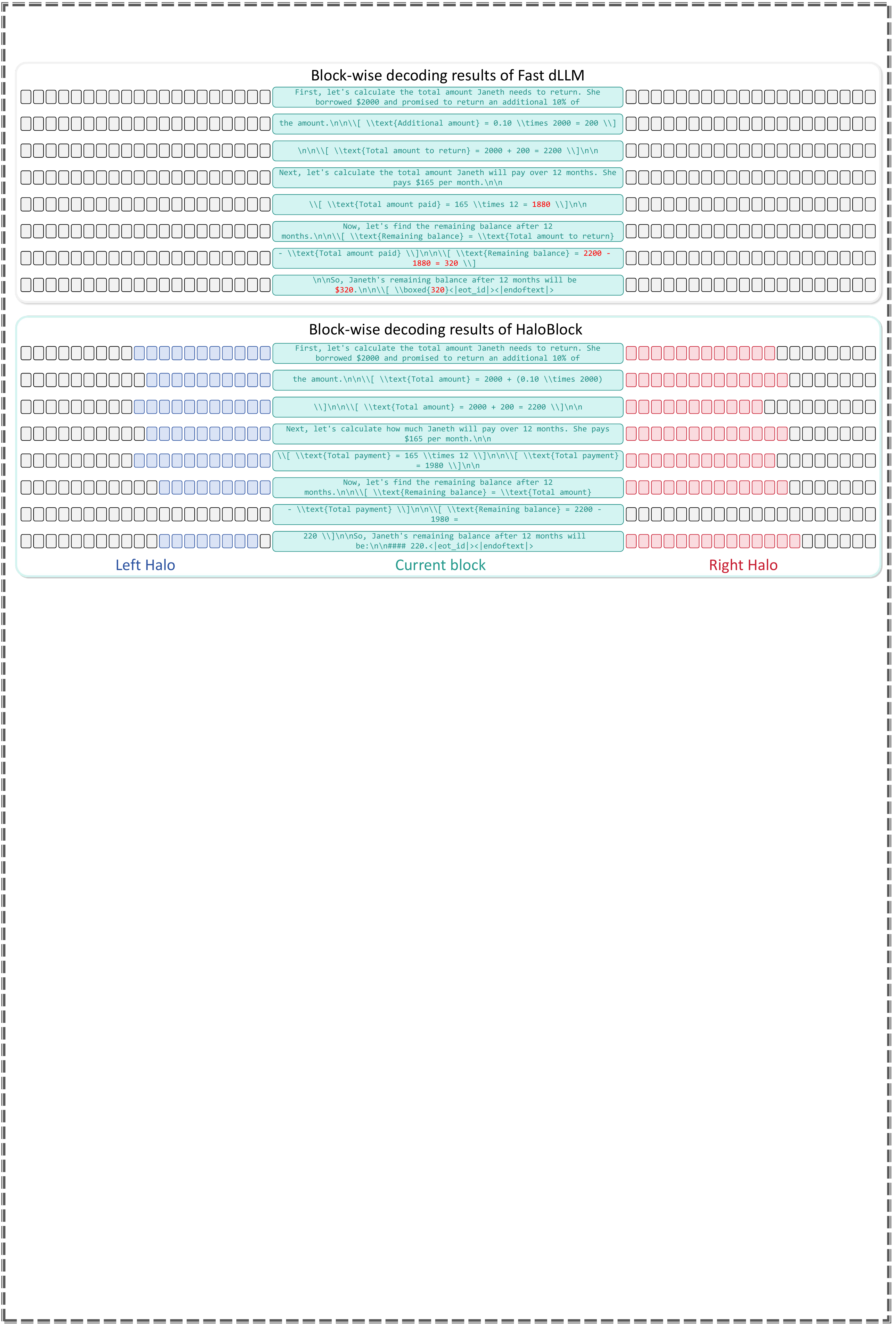}

    \put(1,98){\parbox[t]{0.97\linewidth}{
    Prompt: Janeth borrowed \$2000 and promised to return it with an additional 10\% of the amount. If she is going to pay \$165 a month for 12 months, how much will be Janeth's remaining balance by then?}}
    
    \put(3,55){\footnotesize Note: In the above, the reasoning errors are highlighted in \textcolor{red}{red} and the missing predictions are highlighted in \textbf{bold black}.}
    
    \put(1,52.5){\parbox[t]{0.97\linewidth}{
    \linespread{0.84}\selectfont  
    \textbf{Answer by Fast dLLM}: First, let's calculate the total amount Janeth needs to return. She borrowed \$2000 and promised to return an additional 10\% of the amount.
    \[ \text{Additional amount} = 0.10 \times 2000 = 200 \]
    \[ \text{Total amount to return} = 2000 + 200 = 2200 \]
    Next, let's calculate the total amount Janeth will pay over 12 months. She pays \$165 per month.
    \[ \text{Total amount paid} = \textcolor{red}{165 \times 12 = 1880} \]
    Now, let's find the remaining balance after 12 months.
    \[ \text{Remaining balance} = \text{Total amount to return} - \\\text{Total amount paid} \]
    \[ \text{Remaining balance} = 2200 - 1880 = \textcolor{red}{320} \]
    So, Janeth's remaining balance after 12 months will be \textcolor{red}{\$320}.
    \textcolor{red}{\boxed{320}}}}

    \put(1,27.5){\parbox[t]{0.97\linewidth}{
    \linespread{0.84}\selectfont  
    \textbf{Answer by HaloBlock}: First, let's calculate the total amount Janeth needs to return. She borrowed \$2000 and promised to return an additional 10\% of the amount.
    \[ \text{Total amount} = 2000 + (0.10 \times 2000) \]
    \[ \text{Total amount} = 2000 + 200 = 2200 \]
    Next, let's calculate how much Janeth will pay over 12 months. She pays \$165 per month.
    \[ \text{Total payment} = 165 \times 12 \]
    \[ \text{Total payment} = 1980 \]
    Now, let's find the remaining balance after 12 months.
    \[ \text{Remaining balance} = \text{Total amount}- \text{Total payment} \]
    \[ \text{Remaining balance} = 2200 - 1980 =220 \]
    So, Janeth's remaining balance after 12 months will be:\\
    
    \#\#\#\# 220.
    }}
    \end{overpic}
    \end{minipage}
    \\[2pt]
	\caption{\textbf{Example 2}: Comparison of inference results between the baseline Fast dLLM and HaloBlock. We visualize the dynamic halo length in colors. Reasoning errors are highlighted in \textcolor{red}{red}.}
	\label{fig:example2}
 \vspace{0pt}
\end{figure*}

\begin{figure*}[!t]
    \small
	\centering
	\setlength{\tabcolsep}{1pt}
	\setlength{\abovecaptionskip}{3pt}
    \begin{minipage}{\linewidth}
	\begin{overpic}[width=1\textwidth]{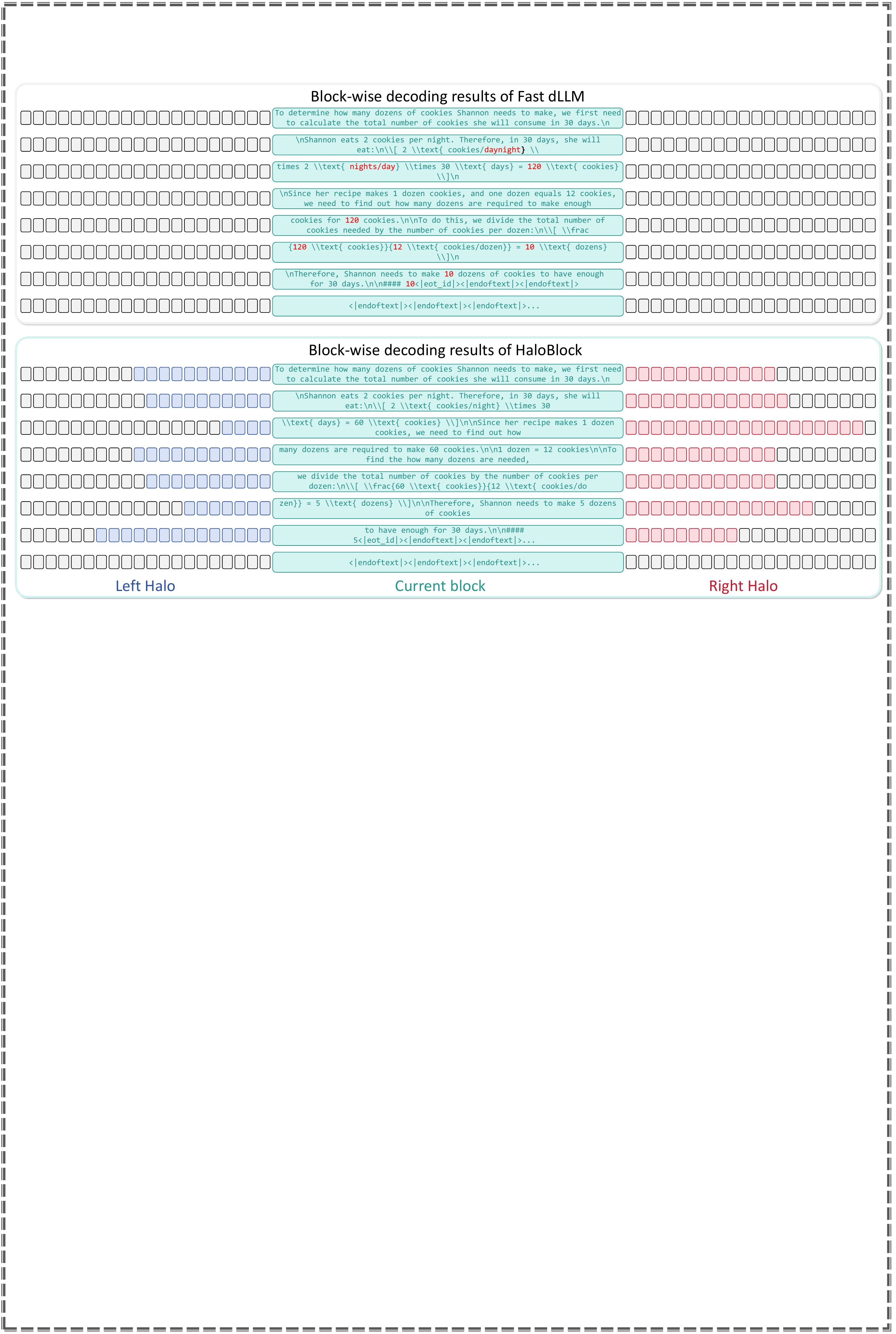}
    \put(1,98){\parbox[t]{0.97\linewidth}{
	Prompt: Shannon makes her own madeleine cookies and eats 2 a night as a treat.  She wants to make enough cookies to last her for 30 days by storing them in the freezer. Her recipe makes 1 dozen madeleine cookies.  How many dozens of cookies will she need to make so she has enough for 30 days?
    }}
    \put(3,53.7){\footnotesize Note: In the above figure, the reasoning errors are highlighted in \textcolor{red}{red} and the missing predictions are highlight in \textbf{bold black}.
    }

    \put(1,51.2){\parbox[t]{0.97\linewidth}{
    \linespread{0.82}\selectfont  
    \textbf{Answer by Fast dLLM}: To determine how many dozens of cookies Shannon needs to make, we first need to calculate the total number of cookies she will consume in 30 days.\\
    
    Shannon eats 2 cookies per night. Therefore, in 30 days, she will eat:
    \[ 2 \text{ cookies/}\textcolor{red}{\text{daynight}} \times 2 \textcolor{red}{\text{ nights/day}} \times 30 \text{ days} = \textcolor{red}{120} \text{ cookies} \]
    Since her recipe makes 1 dozen cookies, and one dozen equals 12 cookies, we need to find out how many dozens are required to make enough cookies for \textcolor{red}{120} cookies.\\
    
    To do this, we divide the total number of cookies needed by the number of cookies per dozen:
    \[ \frac{\textcolor{red}{120} \text{ cookies}}{12 \text{ cookies/dozen}} = \textcolor{red}{10} \text{ dozens} \]
    Therefore, Shannon needs to make \textcolor{red}{10} dozens of cookies to have enough for 30 days.\\
    
    \#\#\#\# 10
    }}

    \put(1,25.4){\parbox[t]{0.97\linewidth}{
    \linespread{0.82}\selectfont  
    \textbf{Answer by HaloBlock}: To determine how many dozens of cookies Shannon needs to make, we first need to calculate the total number of cookies she will consume in 30 days.\\
    
    Shannon eats 2 cookies per night. Therefore, in 30 days, she will eat:
    \[ 2 \text{ cookies/night} \times 30\text{ days} = 60 \text{ cookies} \]
    Since her recipe makes 1 dozen cookies, we need to find out how many dozens are required to make 60 cookies.\\
    
    1 dozen = 12 cookies\\
    
    To find the how many dozens are needed, we divide the total number of cookies by the number of cookies per dozen:
    \[ \frac{60 \text{ cookies}}{12 \text{ cookies/dozen}} = 5 \text{ dozens} \]
    Therefore, Shannon needs to make 5 dozens of cookies to have enough for 30 days.\\
    
    \#\#\#\# 5
    }}
    \end{overpic}
    \end{minipage}
    \\[2pt]
	\caption{\textbf{Example 3}: Comparison of inference results between the baseline Fast dLLM and HaloBlock. We visualize the dynamic halo length in colors. Reasoning errors are highlighted in \textcolor{red}{red}.}
	\label{fig:example3}
 \vspace{0pt}
\end{figure*}

\textbf{Visualization}.
We showcase three mathematical reasoning cases of dLLM inference in \figref{fig:example1}, \figref{fig:example2}, and \figref{fig:example3}.
We also visualize the dynamic halo length for every block.
The changes in the halo region once again exhibit asymmetry, with the right halo being generally longer.
Then, we compare the inference results produced by our HaloBlock and the baseline Fast dLLM \citep{fast-dllm}.
The examples show that HaloBlock benefits from the halo region, which allow the current block to see more contextual information and reduce reasoning errors.

\end{document}